# A Meta-learning based Stacked Regression Approach for Customer Lifetime Value Prediction

Karan Gadgil[1], Sukhpal Singh Gill[1], Ahmed M. Abdelmoniem[1]

[1]School of Electronic Engineering and Computer Science, Queen Mary University of London, UK

*Abstract*— **Companies across the globe are keen on targeting potential high-value customers in an attempt to expand revenue and this could be achieved only by understanding the customers more. Customer Lifetime Value (CLV) is the total monetary value of transactions/purchases made by a customer with the business over an intended period of time and is used as means to estimate future customer interactions. CLV finds application in a number of distinct business domains such as Banking, Insurance, Online-entertainment, Gaming, and E-Commerce. The existing distribution-based and basic (recency, frequency & monetary) based models face a limitation in terms of handling a wide variety of input features. Moreover, the more advanced Deep learning approaches could be superfluous and add an undesirable element of complexity in certain application areas. We, therefore, propose a system which is able to qualify both as effective, and comprehensive yet simple and interpretable. With that in mind, we develop a meta-learning-based stacked regression model which combines the predictions from bagging and boosting models that each is found to perform well individually. Empirical tests have been carried out on an openly available Online Retail dataset to evaluate various models and show the efficacy of the proposed approach.**

**Keywords**— Customer Lifetime Value; Gradient Boosting Machines; Extreme Gradient Boosting.

## I. INTRODUCTION

The key to flourishing businesses lies in understanding the customers using various aspects of their interactions with the businesses. This allows businesses to manage resources in the most targeted fashion. With the recent boom in e-commerce, the majority of customers prefer online shopping over traditional retail because they can compare prices by looking through dozens of websites to locate the best deal, and because of convenience. We live in the Internet age, where technology has advanced significantly. This alteration contributes to the significant rise in retailers in the online retail sector [2]. For any given business the cost to acquire a new customer is more expensive than retaining an existing customer, and therefore online retailers should give more attention to their existing customers. But there will be some customers whose costs of marketing, selling, and servicing can exceed the business's gained profit from them [2].

A customer's total value to a company over the length of their relationship is measured by their customer lifetime value (CLV). In reality, this "worth" could be measured in terms of sales, profits, or other factors that analysts choose. A contractual business is one in which contracts that control the buyer-seller relationship are in place, as the name implies.
The agreement ends when one or both parties decide they no longer wish to be entitled by the contract anymore. The contract eliminates any doubt regarding a person's status as a customer of the company at any given time. This is particularly helpful in churn prediction which forms one element of CLV.

Contrarily, no contract is necessary while doing business in a noncontractual setting because purchases are made as needed. In a continuous environment, purchases can happen at any time. This category includes the bulk of buying circumstances, such as supermarket purchases. In a discrete situation, purchases typically take place intermittently and on a regular basis. Weekly magazine purchases are one example of this [18]. Based on the former analysis, we can categorize our use case in this work to be that of a 'Non-Contractual – Continuous Transactions'.

The business value of a customer is often expressed with CLV which is derived via Equation (1). CLV typically represents the total amount of money (expenditure) a customer is expected to spend in business during their lifetime [15].

$$CLV = \left(\frac{Average\_Sales \times Purchase\_Frequency}{Churn}\right) \times Profit\_Margin \quad \dots\dots\dots\dots\dots\dots\dots\dots\dots\dots\dots\dots\dots\dots\dots\dots\dots\dots\dots\dots \quad (1)$$

Where *Profit_Margin* is based on business context and,

$$Average\_Sales = \frac{Total\_Sales}{Total\_OrderNumber} \quad \dots\dots\dots\dots\dots\dots\dots\dots\dots\dots\dots\dots\dots\dots\dots\dots\dots\dots\dots\dots\dots\dots \quad (1.1)$$

$$Purchase\_Frequency = \frac{Total\_OrderNumber}{Total\_Unique\_Customers} \quad \text{................................................................} \quad (1.2)$$

$$Retention\_Rate = \frac{TotalOrderNumber\_Greater\_than\_1}{Total\_Unique\_Customers} \quad \text{................................................................} \quad (1.3)$$

$$Churn = 1 - Retention\_Rate \quad \text{................................................................} \quad (1.4)$$

Using predictions on the CLV can helps with solving many problems, such as decisions related to segmentation, addressing, retaining, and acquiring customers, or problems concerning a company's long-term value. Over the past few years, numerous methods to investigate how to estimate the CLV have appeared in the literature [3,5,6,15]. The fundamental elements in historical life-time value (LTV) computations originally come from RFM models, which group customers based on recency, frequency, and monetary value (RFM) namely, on how recently and how often they purchased and how much they spent [3,17]. The basic assumption of RFM models is that users with more recent purchases, who purchase more often or who spend larger amounts of money are more likely to purchase again in the future. However, they frequently rely on particular distributional assumptions, which can occasionally result in poor prediction accuracy when the assumptions are not realized.

The success of various machine learning (ML) approaches in recent years for numerous real-world applications has generated interest in using these techniques for the CLV prediction problem and the related but less challenging churn prediction problem [13]. Moreover, CLV prediction can be projected as a time series forecasting problem which offers an alternate perspective on this problem [15]. To be more precise, we can attempt to use historical transaction data to estimate future time slots for the following few days or weeks, and then combine these predictions to determine the overall CLV for that period of time. The benefit of time series modelling is that the sequential nature of the data is kept as opposed to methods that explicitly model the CLV, such as by summing expected profits per time step [13,15].

In this work, we aim to build a robust predictive model for CLV. To this end, we propose to leverage a meta-learning-based stacked regression model which combines the predictions from different well-performing models. To be able to apply time-series data with the goal of implementing a predictive model, the input dataset needs to be pre-processed in a way which involves organizing the data into dependent and independent variables and dealing with null values. Additionally, feature selection needs to be performed to pick the most relevant features which contribute towards prediction. Multiple models have been implemented and compared against our proposed ensemble-based meta-model in terms of the RMSE & MAE as our primary comparison metric given our time-series regression problem.

*A. Motivation*

The CLV can tell a business owner a great deal of information about the business and finds application in a number of distinct business domains like banking, insurance, online entertainment (such as over-the-top platforms), gaming, and e-commerce. In the following, we present some of the most important ways the lifetime value of a customer can be of use to any business that is influenced by its customer interaction:

> 1. **Budget allocation**: The CLV offers essential information for budget allocation. A business may more effectively plan where to invest in the expansion of your business if they are aware of how much the typical customer spends and thereby maximize the return on advertising, brand-building, and marketing plan investments by concentrating on the most engaged and devoted clients.
> 2. **Income projection**: Accurate CLV data can be used to estimate future revenue. This allows for the business to bank on future sales from loyal consumers, which will result in a steady income.
> 3. **Customer information**: The CLV has the potential to provide important details about the people who are purchasing goods from a business and this can guide decisions on how to allocate resources to increase customer loyalty. Additionally, it enables more effective client segmentation, thus allowing to divide of the customer base into several groups, from the irregular to the periodic, for focused marketing.
> 4. **Customer contentment**: By understanding the customers' behaviour, a business can plan ahead to raise customer satisfaction, decrease churn rates (the rate at which customers discontinue using your product or service), and improve customer retention.

*B. Contributions*
- We study and analyze the problem of CLV prediction and present the prominent existing methods used for this purpose.
- We propose a Meta-learning based Stacked-Regression approach to tackle this problem which presents a novel solution to the problem.
- Using a common retail dataset, we evaluate the proposed approach and compare them with the existing methods to show its effectiveness.

- Our results show that the proposed approach shows superior performance and can accurately predict the CLV with low errors.

## C. Paper Organization

The remainder of the paper has been arranged as follows. Section II will review the work related to CLV prediction. Section III will describe the methodology consisting of dataset description, data pre-processing, experimentation and results related to the models tested. A discussion of the conclusions and a perspective for future works conclude this paper under section IV.

## II. RELATED WORK

### A. Negative Binomial Distribution (NBD) Model

A family of more complex probabilistic approaches have been proposed in the research literature, and like some of the Markov Chain (MC)-based approaches, they are motivated by the notion that the CLV prediction process can be divided into two components. To achieve this, the following are forecasting challenges to be dealt with: 1) the first issue is determining whether a buyer will make another purchase or not; 2) the second issue has to do with the number of orders and the anticipated profit.

The unique concept behind these models, in contrast to MC approaches, is that each step in the process is based on a separate distributional assumption, i.e., every customer's purchase process is viewed as a manifestation of a certain probability distribution. NBD-based strategies have the advantage of being logical and based on well-established principles. These methods perform best when the specific distributional assumptions are true or nearly true, and when the CLV is not significantly impacted by other hidden variables. However, in reality, these presumptions are not always true, which reduces the prediction effectiveness of these models. Furthermore, these models do not consider other predictor variables or the fact that the data are time series.

### B. Bagging method – Random forest

In [2], the use of random forest (RF) which is a supervised machine learning algorithm has been discussed for the purpose of customer classification based on a customer's spending value. This approach is well suited to a more general level of use case granularity of customer segmentation.

In a regression problem, the highest and lowest labels in the training data serve as a boundary for the range of predictions that a Random Forest model can produce [23]. When the range and/or distribution of the training and prediction inputs change, this behaviour could have varying impacts. Since Random Forest methods can't extrapolate, it is challenging to handle the common phenomenon known as covariate shift [11].

A new feature of the system proposed in [6] uses the customer's view history to obtain clickstream data which is of a sequential nature having data related to the fashion industry. In circumstances with sparse data, i.e. zero value records, researchers frequently employ embedding representations (which typically finds application in NLP) rather than the raw sequential data directly.

The authors in [6] learn embeddings using logged item view events in the context of CLV prediction to find clients with related interests. The CLV prediction system in [5] employs an RF model that allows accounting for a wide range of aspects of customer engagement with the platform. Moreover, it showed to be evidently promising in terms of being able to handle a more feature-rich dataset and offer better performance in terms of prediction.

After understanding both the merits and demerits of a Random Forest model, we intend to use this as one of the base models in combination with other base models such as XGBoost to help compensate for any drawbacks or inaccuracies in RF models.

### C. Boosting Method – Gradient Boosting method XGBoost

Regression trees serve as the weak learners when utilising gradient boosting for regression, and each one of them associates each input data point with a leaf that holds a continuous score. For this purpose, a convex loss function (based on the difference between the predicted and target outputs) and a penalty term for model complexity are used. XGBoost in particular minimises a regularised (L1 and L2) objective function [19]. Figure 1 is an illustration of how gradient tree boosting works.

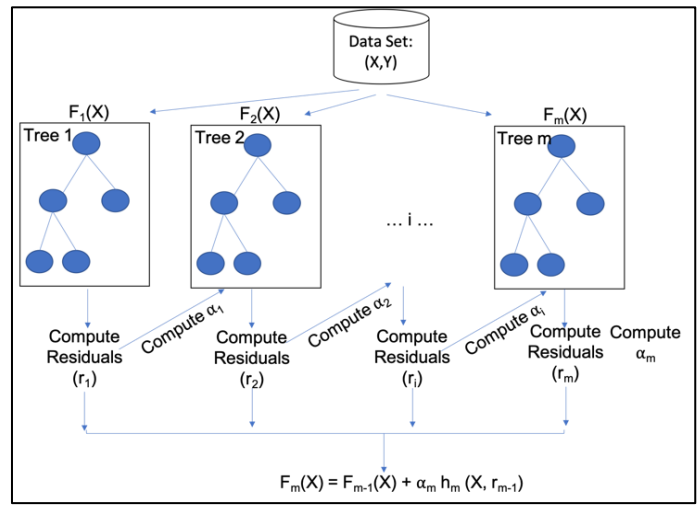

*Figure 1. Architectural Design of XGBoost*

In [7], the authors compare the SVM with XGBoost, a special type of gradient boosting machine learning algorithm that works by combining hundreds of simple trees with a low accuracy with the target of building a more accurate model. The XGBoost model in [7] adopts a tree model approach as a booster out of several options for solving a stock selection regression problem. An essential feature of such a gradient-boosting algorithm is that it significantly reduces over-fitting problems commonly seen in various classes of applications with the help of a regularization term and provides abilities for achieving distributed and parallel computing [7]. The gradient boosting tree ensemble method was also applied and found to have a better prediction accuracy.

Machine learning is frequently criticised for functioning like a "black box" where we input data on one side and get the result on the other. Although the responses are frequently quite accurate, the model doesn't explain how it came up with the forecasts. This is somewhat true, however there are approaches to attempt and figure out how a model "thinks," like the Locally Interpretable Model-agnostic Explainer (LIME) [8]. Additionally, there exist other primitive means such as discovering feature importance and subsequently using the ones rendered to be more decisive and contribute the most towards the predictions. LIME by learning a linear regression around the prediction, which is an understandable model, tries to explain model predictions [8]. Since we are concerned with interpretability to a reasonable degree, employing such a method is also likely to enhance the interpretability of the model's outcomes.

Therefore, we aim to use a boosting ensemble model as one of the base models so as to conform to the required heterogeneity for taking a stacking approach.

### D. Deep Learning Models

In the realm of gaming data science, DNNs have been applied to the simulation of in-game events [3] as well as the forecasting of churn and purchases. The authors in [3] aim to predict the purchases a player will make from the day of the prediction until they exit the game, which could be anywhere between a few days and a few years. While focused on forecasting the annual total purchases from player activity during the first seven days of the game. An input layer, numerous hidden layers, and an output layer make up a deep multilayer perceptron. The input of the input layer consists of features (user activity logs), while the output of the output layer is the prediction result (LTV). Neurons with nonlinear activation functions generate layers that are connected. The neural network is optimized through a number of iterations, or epochs, during the learning process.

The proposed system in [9] aims to leverage a deep learning approach (typically used in image classification) in the context of telecommunications for churn prediction which is a related aspect of CLV. The problem of customer churn has been attempted to be solved using both methods, i.e. supervised and unsupervised [9]. The authors also tried to customize the system to each customer by developing a two-dimensional array having rows that represent various means of communication and columns that represent days of the year (e.g., text, call etc.).

A customized CLV prediction employing gated sequential multi-task learning in the context of online gaming is presented in [10] which focuses on CLV factors, such as customer churn, payment/revenue in isolation, as well as their correlation with one another. The authors introduce an interesting approach in [10] where they also observe and draw patterns based on the individual player behavior to ascertain their effects on the churn and payment individually which unveils fascinating findings such as which players tend to use up their tokens/resources before churning and high-paying players who engage more in competitive gameplays. Additionally, they also explore social behavior influenced by players from each other. As their in-game activity is influenced by the nearby players, Churn players instinctively group together to establish several little local groups. A player is more likely to soon churn if the majority of their friends around them do so as well. There is a good chance that a player will stay active if the majority of their friends are also active. Experiments have been conducted on three real-world datasets, two of which include mobile games of different genres and a publicly available advertising dataset made available by Alibaba [10]. In the experiments, the number of parameters and time is directly proportional to the accuracy achieved and therefore it affords a high level of computational complexity and is highly likely to compromise on the interpretability factor. For the aforementioned reasons, we intend to restrict our proposed solution to only leveraging a machine learning model. However, we note that any behavioral and social patterns discussed above if available could be definitely incorporated and are likely to lead to better performance.

### III. METHODOLOGY

*A. Our Proposed model – Architecture*

We propose a novel approach towards customer lifetime value estimation with the aim of achieving improved performance over existing methods. Given the class of our problem which is a time-series based regression problem, the model makes use of a stacking-based approach that leverages a set of base models consisting of multiple regressors: **RandomForest regressor, XGBoost regressor and an elasticNet regressor**. These base models are trained on the input feature set. The predictions from these models are further fed as input to a meta-model such as a linear regressor (elasticNet) along with the original of inputs to create the final predictions of the proposed model.

The stacking approach forms a component in the system outlined in [13] and we find this to be a valuable approach to further build upon. However, in this work we utilize it in the context of CLV predictions using our own distinct combination of level-1 base models based on the stacking guidelines in [12].

Figure 2 shows a high-level architecture of the proposed system. A stacking-based model works with 2-levels as depicted where level-1 is trained on multiple models on the same dataset followed by level-2 which can be achieved by means of averaging or a meta-model. In our case, we implement stacking, also termed blending, using a meta-model approach. The meta-learning approach works by finding the best way of combining the predictions from ensemble members or the base models in level-1.

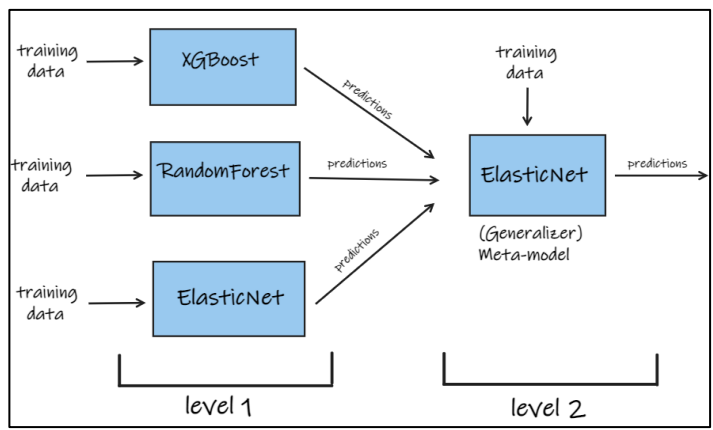

*Figure 2. High Level Model Architecture*

The base models' predictions from training data are used to train the meta-model. To put it another way, data that wasn't used to train the base models are provided to the base models, and then those predictions, along with the expected outputs, serve as the input and output pairs for the training dataset that was used to build the meta-model. Desirable practices that were suggested in [12] for using a stacking approach:
1. Heterogeneous models as base models.
2. Use a linear regressor as a meta-model or generalizer.
3. The base models used should have skill on the problem at hand we are trying to solve but skilful in different ways. In simple terms, their predictions should be un-correlated and use different internal representations of the training data.

Our proposed approach correctly adheres to all the prescribed practices of designing a proper stacking ensemble model. The base models used all follow distinct approaches and are capable of training data representation that is non-overlapping. Additionally, the final generalizer or the meta-model used is also a simple linear regressor.

In order to prepare the input data for the level-2 regressor, the StackingCVRegressor that we utilise extends the conventional stacking approach (implemented as StackingRegressor)[14]. The first-level regressors are fitted to the same training set that is used to produce the inputs for the second-level regressor in the traditional stacking approach, which could result in overfitting. StackingCVRegressor, on the other hand, makes advantage of the idea of out-of-fold predictions. We refer to Figure 3 which explains the full process. First, the dataset is divided into k folds, and in k subsequent rounds, k-1 folds are used to fit the first level regressor [14]. The last 1 subset that was not used for model fitting in each iteration receives the first-level regressors in each round. The generated predictions are then stacked and fed into the second-level regressor as input data.

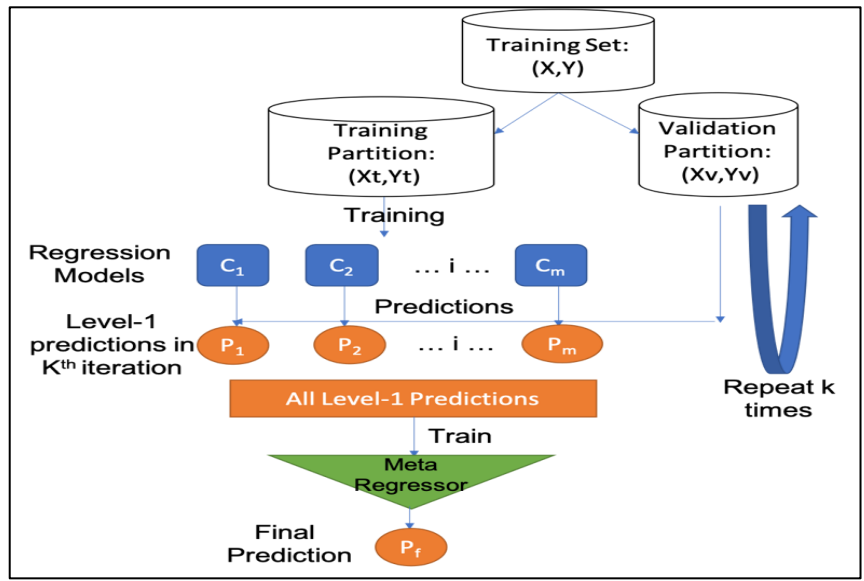

Figure 3. Cross Validation StackingCVRegressor

B.   *Dataset Description*

The '*Online Retail II*' [22] dataset which was made available by UCI. It consists of transactions made by a UK-based, registered, and non-store online retailer between Dec1, 2009, and Dec 30, 2011. Figure 4 shows a smaple view of the dataset and its attributes.

| Invoice | StockCode | Description | Quantity | InvoiceDate | Price | Customer ID | Country |
|---|---|---|---|---|---|---|---|
| 489434 | 85048 | M CHRISTMAS GLASS BALL 20 LIGHTS | 12 | 01-12-2009 07:45 | 6.95 | 13085 | United Kingdom |
| 489434 | 79323P | PINK CHERRY LIGHTS | 12 | 01-12-2009 07:45 | 6.75 | 13085 | United Kingdom |
| 489434 | 79323W | WHITE CHERRY LIGHTS | 12 | 01-12-2009 07:45 | 6.75 | 13085 | United Kingdom |
| 489434 | 22041 | RECORD FRAME 7" SINGLE SIZE | 48 | 01-12-2009 07:45 | 2.1 | 13085 | United Kingdom |
| 489434 | 21232 | STRAWBERRY CERAMIC TRINKET BOX | 24 | 01-12-2009 07:45 | 1.25 | 13085 | United Kingdom |
| 489434 | 22064 | PINK DOUGHNUT TRINKET POT | 24 | 01-12-2009 07:45 | 1.65 | 13085 | United Kingdom |
| 489434 | 21871 | SAVE THE PLANET MUG | 24 | 01-12-2009 07:45 | 1.25 | 13085 | United Kingdom |
| 489434 | 21523 | FONT HOME SWEET HOME DOORMAT | 10 | 01-12-2009 07:45 | 5.95 | 13085 | United Kingdom |
| 489435 | 22350 | CAT BOWL | 12 | 01-12-2009 07:46 | 2.55 | 13085 | United Kingdom |
| 489435 | 22349 | DOG BOWL , CHASING BALL DESIGN | 12 | 01-12-2009 07:46 | 3.75 | 13085 | United Kingdom |

Figure 4. A Sample from Original Dataset

The company has a large number of wholesalers as clients. Table 1 shows the basic analysis obtained from the dataset. While Figure 5. shows the probability (frequency) distribution function of the customer purchases.

Table 1. Basic Analysis of the Dataset

| Number of customers: 5850 |
|---|
| Number of transactions: 397,924 |
| Date: 2009-12-01 to 2011-12-09 |
| Average CLV: 2053.79, median CLV: 674.45 |

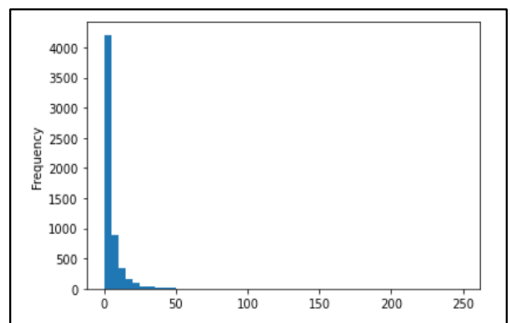

Figure 5. Customer Purchase Frequency Histogram

*C.   Data Cleaning and feature engineering*

The time allocated to data cleaning and feature engineering process comprises the biggest chunk of the entire data science and solution-building process which we corroborate based on our experience. In this process, we have carried out the primary analysis and observation discussion to obtain even more impactful and quality data to be input into the set of models that we intend to test and evaluate. Different model types could be capable of handling input data with varying feature complexities. For example, a basic RFM (Recency, Frequency, Monetary) based model is less likely to be able to handle more domain-specific features. Data consisting of cancellation records have been isolated into a separate column and further dropped as it does not contribute to the prediction performance. Moreover, records with unit prices equal to zero have been removed. A new feature named revenue has been added obtained as a result of multiplying the corresponding 'quantity' and 'price' values. The dataset obtained after de-noising was subsequently grouped based on the customer id and invoice count and revenue sum as can be seen in Table 2.

*Table 2. Customer-wise Invoice and Revenue*

| CustomerID | Invoice | Revenue |
|---|---|---|
| 12346 | 34 | 77556.46 |
| 12347 | 242 | 5408.5 |
| 12348 | 51 | 2019.4 |
| 12349 | 175 | 4428.69 |
| 12350 | 17 | 334.4 |

*Table 3. Feature Engineered data-frame*

| CustomerID | latetime | earlytime | freq | freq_3m | target |
|---|---|---|---|---|---|
| 14911 | 5 | 638 | 203 | 33 | 47 |
| 12748 | 1 | 635 | 159 | 29 | 37 |
| 17841 | 3 | 637 | 154 | 29 | 35 |
| 15311 | 12 | 638 | 168 | 20 | 25 |
| 14606 | 3 | 636 | 157 | 19 | 22 |

Table 3 shows the actual data-frame obtained after cleaning and feature engineering which is further used to train the machine learning models where we have a set of predictors viz. 'latetime', 'earlytime', 'freq', 'freq3m' and the 'target' variable which indicates the 'number of transactions' of each customer_id for the succeeding 3 months.

## IV.   EXPERIMENTAL RESULTS

The models used to compare against each other generally outperform the basic BG/NBD model. Albeit our primary purpose is to ascertain how better are our more sophisticated machine learning models collectively against the baseline BG/NBD. But, more importantly, we also aim to understand which among the set of machine learning models is the best performing in terms of accuracy of estimating customer value for a given future time period. Since the dataset does not possess other personal information data which could be informative such as customer age, we have not been able to consider the customer churn predictability. Below we detail how the metrics used for feature importance comparison.

**Feature Importance** - Indicating the relative importance of each feature while producing a prediction, feature importance refers to a set of strategies for assigning scores to input features to a predictive model. For problems involving the prediction of a numerical value (called regression problems) and problems involving the prediction of a class label (called classification problems), feature significance scores can be computed. The following are the key indicators that could be used for measuring feature importance:
    **Gain** -   refers to the average gain across all splits when the feature is used.
    **Weight** - refers to the frequency with which a feature is utilized to distribute the data among all branches.
    **Cover** - refers to the feature's overall average coverage across all splits. (proportion yes/no).
    **Total gain** - is the overall profit from all splits where the feature is used.
    **Total cover** - represents the feature's overall coverage across all splits.

For our regression problem through the modules provided by scikit-learn for the models LightGBM and XGBoost, we analyse which input features contribute the most in terms of the '**weight**' and '**gain**' indicators.

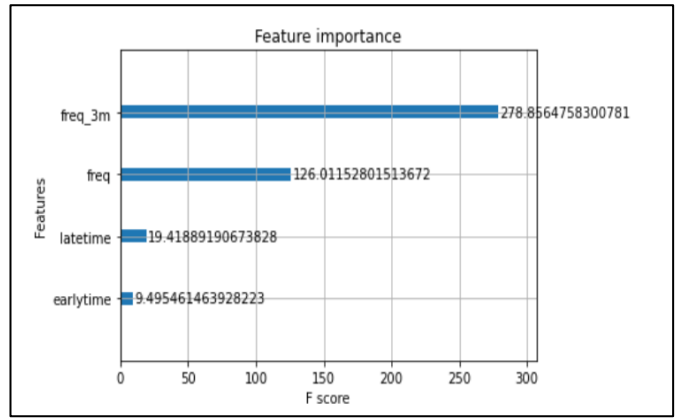

*Figure 6. (XGBoost) Feature Importance Type – Gain*

Figures 6 and 7 illustrate the feature importance for an XGBoost model based on '**Gain**' and '**Weight**' showing that the most important features are different in each and they are *freq_3m* and *latetime*, respectively. In our implementation, we have generated similar graphs for other models like LightGBM and RandomForest.

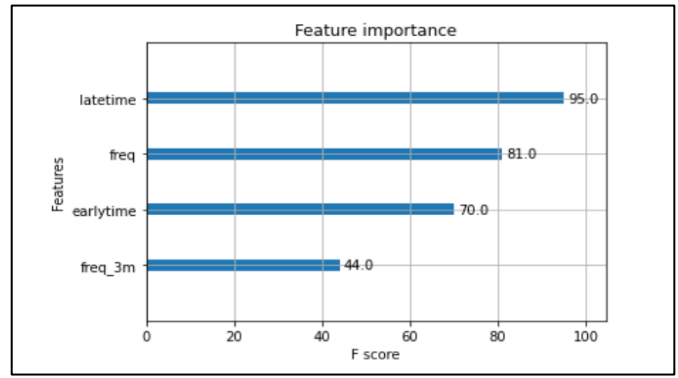

*Figure 7. (XGBoost) Feature Importance Type – Weight*

a) *Configuration settings*

*Table 4. Model Configuration Settings*

| Model | Parameters/Configuration |
|---|---|
| LightGBM | n_estimators': 200, max_depth': 2, 'learning_rate': 0.02 |
| XGBoost | n_estimators=10 |
| RandomForest | n_estimators = 200, max_features = 'sqrt', max_depth = 50 |
| Stacked Regressor | meta_regressor= regr, use_features_in_secondary=True |

Multiple parameters relating to each model exist but some essential features worth exploring are highlighted in Table 4 above. The 'use_features_in_secondary' value in Stacked Regressor has been set to 'True' as it lets the meta regressor take the original training set as an input in addition to the predictions from level 1 models. Setting this value to False instead would let only take predictions of level 1 models as input which gives a poor score for the overall model.

b) *Evaluation Metrics*

Our evaluation metrics for the predictions are the **mean absolute error (MAE)** and the **root mean square error (RMSE)**. Both quantify the discrepancy between a model's anticipated CLVs and the test set's actual values. The RMSE measure penalizes bigger deviations more severely, which is the difference. If it is particularly critical to identify high-value consumers, this can be beneficial in application settings where larger errors pose a proportionally higher risk to the firm. The MAE's benefit is that it may be used to calculate the average error in terms of money [13].

It is worth noting that there are additional metrics besides the MAE or RMSE frequently employed in the research literature on time series regression, such as the mean absolute value percentage error (MAPE). Such percentage-based measurements in our situation, where we deal with a particular application domain, are less informative than those that operate on the absolute monetary values. For instance, take two customers, one whose CLV is predicted to be £5 but is actually £10, and another whose CLV is expected to be £50 but is actually £100. Although the absolute numbers are substantially bigger in the second scenario, the relative error would be the same, and the company would be at greater danger or loss if the estimate was incorrect [13].

*c)    Results*

Table 5. Results Comparison Table

| Method | OnlineRetail RMSE | OnlineRetail MAE |
| --- | --- | --- |
| BG/NBD | 1.62 | 0.9 |
| LightGBM | 1.95 | 0.88 |
| DNN | 1.53 | 0.83 |
| RandomForest | 1.44 | 0.87 |
| XGBoost | 1.34 | 0.83 |
| **Stacked Regressor (Proposed )** | **1.37** | **0.82** |

The results in terms of the chosen evaluation metrics are shown in Table 5. The models tested have been arranged in descending order of their scores and a lower value across both metrics is desirable. Our proposed model generally outperforms the selected baseline model as well as other models having a machine learning approach in both metrics by achieving the lowest MAE value and the second smallest RMSE value.

BG/NBD which we treat as our baseline model is more fundamental in that it takes into account the underlying distributions. The LightGBM model performs slightly better by getting a lower RMSE value although this cannot be considered significant against the baseline BG/NBG. Deep learning approaches are becoming increasingly common and are performing significantly well. Although extensively exploring the deep learning approach is not our primary focus, we still intend to provide a sense of how well a machine learning approach is in juxtaposition with DNN. To that end, it was found that the stacked regressor model has been found to do better than a DNN tuned to an optimum configuration setting with our dataset.

The bagging and boosting techniques viz. RandomForest and XGBoost handle this problem well however the RMSE score achieved is lower for XGBoost regressor model. The Stacked regressor (proposed model) was able to achieve a slightly lower RMSE and MAE as expected based on an intuitive understanding of the inherent concept.

These results suggest that our proposed stacking method achieves the best results and it provides for more robust ML systems with higher levels of interpretability.

*d)    Discussion*

The availability of a number of possibly informative and helpful attributes/features in the context of CLV prediction discussed ahead if present in the dataset could lead to improved results:

1) **Bank-Holidays**, signs of the run-up to and after of holidays, unique seasons, and global trends: The average profit over all consumers might be impacted by holidays and other temporal occurrences. Additionally, important business actions like reducing shipping times can alter the overall mean.

2) **Clickstream Information Combined with Behavioral Data from the Website and Outside Sources**: These kinds of fine-grained data, such as customer reviews, discounts applied, and returned goods, might be employed as extra predictors in future models.

3) **Information about marketing promotions**: Understanding current and upcoming marketing initiatives frequently has a direct impact on sales. It appears promising to include such information in the CLV prediction model.

4) **Not in Stock or Available Information on products that customers have previously preferred or may choose in the future**: When we are aware that a customer's preferred things are not accessible, we might anticipate that their overall spending may be less.

## V. CONCLUSION AND FUTURE WORKS

In this paper, we have introduced a rare meta-learning based stacking approach with a new underlying combination of bagging and boosting methods previously found effective individually in the context of CLV which gives a promising result by attaining a lower value of RMSE and MAE values. This proves to be more comprehensive in terms of flexibility towards being able to accommodate more features than primitive techniques as well as match up to the performance of DNNs. This has been evaluated using metrics viz. Root Mean Squared Error (RMSE) and Mean Absolute Error (MAE) are considered to be relevant to the time series regression problem.

The above discussed additional features and possible handcrafted features resulting from them would add more certainty and confidence in the performance of our proposed system. Testing the proposed model in a different context or a different aspect of CLV would help realize its versatility and applicability for other classes of problems.

One promising addition lies in considering the temporal patterns consisting of seasonality and trends with the help of an encoder-decoder RNN as an additional component of our current system without the need for manual feature engineering [13]. Inputting the data values as embeddings [6] is likely to prove another promising methodology in terms of data pre-processing and feature engineering.

## VI. ACKNOLWDGEMENTS

## REFERENCES

1. Pfeifer and Robert L. Carraway. 2000, "Modeling customer relationships as Markov chains" Journal of Interactive Marketing 14, 2 (2000), 43-55.

2. T. T. Win and K. S. Bo, "Predicting Customer Class using Customer Lifetime Value with Random Forest Algorithm," 2020 International Conference on Advanced Information Technologies (ICAIT), 2020, pp. 236-241

3. Pei Pei Chen, Anna Guitart, Ana Fernández del Río, África Periáñez, "Customer Lifetime Value in Video Games Using Deep Learning and Parametric Models", arXiv:1811.12799 (2018)

4. Chun-Yao Huang, "To model, or not to model: Forecasting for customer prioritization", International Journal of Forecasting, Volume 28, Issue 2, 2012, Pages 497-506.

5. Ali Vanderveld, Addhyan Pandey, Angela Han, and Rajesh Parekh. 2016. "An Engagement-Based Customer Lifetime Value System for E-commerce", In Proceedings of the 22nd ACM SIGKDD International Conference on Knowledge Discovery and Data Mining (KDD '16).

6. Benjamin Paul Chamberlain, Angelo Cardoso, Chak H. Liu, Roberto Pagliari, and Marc Peter Deisenroth. 2017. "Customer lifetime value prediction using embeddings". In Proceedings of the 23rd ACM SIGKDD International Conference on Knowledge Discovery and Data Mining. 1753–1762

7. Haoyue Liu. 2020. "Stock Selection Strategy Based on Support Vector Machine and eXtreme Gradient Boosting Methods". In 2020 the 4th International Conference on Big Data Research (ICBDR'20) (ICBDR 2020).

8. H. Ma, X. Yang, J. Mao and H. Zheng, "The Energy Efficiency Prediction Method Based on Gradient Boosting Regression Tree," 2018 2nd IEEE Conference on Energy Internet and Energy System Integration (EI2), 2018, pp. 1-9

9. Artit Wangperawong, Cyrille Brun, Olav Laudy, Rujikorn Pavasuthipaisit, "Churn analysis using deep convolutional neural networks and autoencoders", arXiv:1604.05377 (2016)

10. Shiwei Zhao, Runze Wu, Jianrong Tao, Manhu Qu, Minghao Zhao, Changjie Fan, and Hongke Zhao. 2022. "PerCLTV: A General System for Personalized Customer Lifetime Value Prediction in Online Games". ACM Trans. Inf. Syst. Just Accepted (April 2022)

11. Ben Thompson, "A limitation of Random Forest Regression", Towards Data Science (2019)

12. Jason Brownlee, "How to Implement Stacked Generalization (Stacking) From Scratch With Python", machine learning mastery (2016)

13. Josef Bauer and Dietmar Jannach. 2021. "Improved Customer Lifetime Value Prediction With Sequence-To-Sequence Learning and Feature-Based Models", ACM Trans. Knowl. Discov. Data 15, 5, Article 80 (October 2021)

14. "StackingCVRegressor: stacking with cross-validation for regression". mlxtend user guide, URL: http://rasbt.github.io/mlxtend/user_guide/classifier/StackingCVClassifier/

15. P Jasek, L Vrana, L Sperkova, Z Smutny, M Kobulsky, "Modeling and application of customer lifetime value in online retail", Informatics, MDPI Publishing, 2018

16. Fader, P. S., Hardie, B. G., & Lee, K. (2005). "Counting Your Customers" the Easy Way: An Alternative to the Pareto/NBD Model. Marketing Science, 24 (2), 275-284.

17. Fader PS, Hardie BGS, Lee KL. "RFM and CLV: Using Iso-Value Curves for Customer Base" Analysis. Journal of Marketing Research. 2005;42(4):415-430.

18. Adam Ramshaw "Calculating and Using Customer Lifetime Value". Genroe, URL: https://www.genroe.com/blog/customer-lifetime-value/15199

19. "How XGBoost Works". Amazon SageMaker Developer Guide, URL: https://docs.aws.amazon.com/pdfs/sagemaker/latest/dg/sagemaker-dg.pdf#xgboost-HowItWorks

20. Vineet Kumar and Werner Reinartz. 2018. "Customer Relationship Management: Concept, Strategy, and Tools'. Springer (2018)

21. Tianqi Chen and Carlos Guestrin. 2016. XGBoost: A scalable tree boosting system. In Proceedings of the 22nd ACM SIGKDD International Conference on Knowledge Discovery and Data Mining (KDD'16). ACM (2016).

22. Online Retail II Data Set. URL: https://archive.ics.uci.edu/ml/datasets/Online+Retail+II

23. Sukhpal Singh Gill, and Rupinder Kaur. "ChatGPT: Vision and challenges." Internet of Things and Cyber-Physical Systems 3 (2023): 262-271.